%% file: Template_ISBI_latex.tex
\documentclass{article}
\usepackage{spconf,amsmath,graphicx}
\usepackage{amsfonts}
\usepackage[flushleft]{threeparttable}
\usepackage{booktabs}
\usepackage{tabularx}
\usepackage{multirow}
\usepackage{subcaption}
\usepackage[table]{xcolor}
\usepackage{bm}

\usepackage{xcolor}

\usepackage{enumitem}
\setlist{nosep, leftmargin=14pt}

\usepackage{hyperref}

\DeclareMathOperator*{\argmin}{arg\,min}

\title{Covariance Descriptors Meet General Vision Encoders:\\Riemannian Deep Learning for Medical Image Classification}

\name{Josef Mayr*$^{1}$, Anna Reithmeir*$^{1,2,3}$, Maxime Di Folco$^{3}$, Julia A. Schnabel$^{1,2,3,4}$}
\address{
$^{1}$School of Computation, Information and Technology, Technical University of
Munich, Germany\\
$^{2}$Munich Center for Machine Learning (MCML), Germany\\
$^{3}$Institute of Machine Learning in Biomedical Imaging, Helmholtz Munich, Germany\\
$^{4}$School of Biomedical Engineering and Imaging Sciences, King’s College London,
UK
\thanks{
This work has been submitted to the IEEE International Symposium on Biomedical Imaging (ISBI) 2026 for possible publication. 
Copyright may be transferred without notice, after which this version may no longer be accessible.
}
}

\begin{document}

\maketitle

\begin{abstract}
Covariance descriptors capture second-order statistics of image features. They have shown strong performance in general computer vision tasks, but remain underexplored in medical imaging. 
We investigate their effectiveness for both conventional and learning-based medical image classification, with a particular focus on SPDNet, a classification network specifically designed for symmetric positive definite (SPD) matrices.
We propose constructing covariance descriptors from features extracted by pre-trained general vision encoders (GVEs) and comparing them with handcrafted descriptors.
Two GVEs -- DINOv2 and MedSAM -- are evaluated across eleven binary and multi-class datasets from the MedMNSIT benchmark.
Our results show that covariance descriptors derived from GVE features consistently outperform those derived from handcrafted features. Moreover, SPDNet yields superior performance to state-of-the-art methods when combined with DINOv2 features.
Our findings highlight the potential of combining covariance descriptors with powerful pretrained vision encoders for medical image analysis.
\end{abstract}


\section{Introduction}
\label{sec:intro}

In general computer vision, covariance descriptors have demonstrated strong performance across diverse tasks, including object detection \cite{tuzel2006region}, texture classification \cite{tuzel2006region}, shape matching \cite{Tabia_2014_CVPR}, and action recognition \cite{sanin2013spatio}. 
These descriptors represent images as symmetric positive definite (SPD) matrices computed from the covariances of extracted image features.
Despite their potential, covariance descriptors remain largely unexplored in medical image analysis, including medical image classification.

Recently, large pre-trained general vision encoders (GVEs) based on Vision Transformers (ViTs) \cite{dosovitskiy2020vit} have emerged as powerful tools for capturing rich and transferable visual representations.
Prior work on covariance descriptors has relied primarily on handcrafted features, and their application in medical imaging has been limited \cite{Khan2015, Stanitsas2016, Cirujeda2015_3D, Ahmadi2024}. 
In particular, covariance descriptors derived from learned representations, such as those provided by GVEs, have not yet been investigated.
A central challenge in working with covariance descriptors is their non-Euclidean geometry: The set of SPD matrices forms a Riemannian manifold, making standard Euclidean-based methods suboptimal. Algorithms must therefore account for the manifold topology, for instance, through the use of Riemannian metrics. Manifold-aware deep learning methods are particularly well-suited for this but have not been explored in the context of covariance-based medical image classification yet.

\begin{figure}[t]
    \centering
    \includegraphics[width=0.95\linewidth]{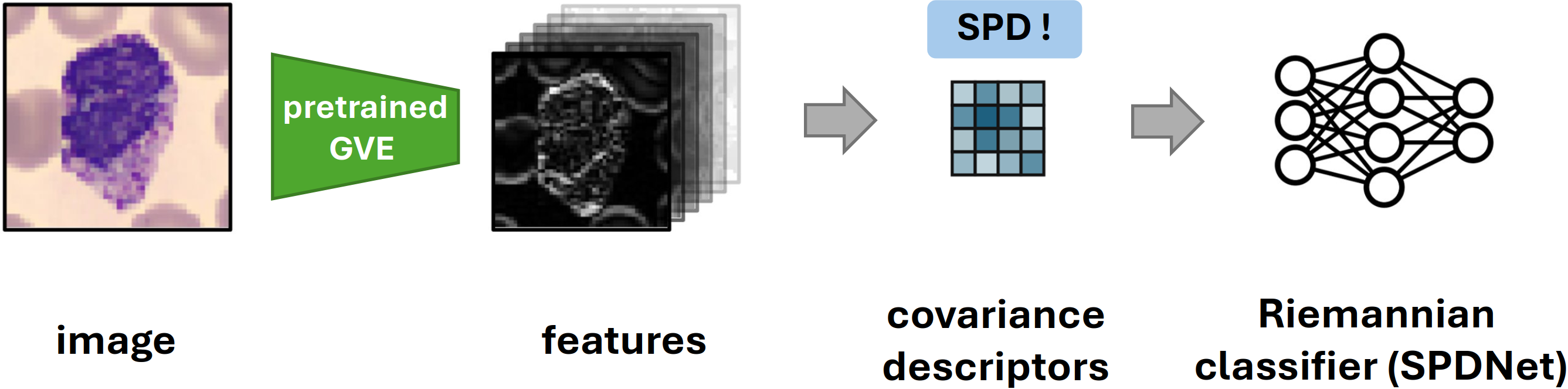}
    \caption{Overview of proposed method. First, general vision encoder (GVE) features are extracted for an image. Then, the covariance descriptors are computed, which are used as input to a learning-based Riemannian classifier. 
    }
    \label{fig:overview}
\end{figure}

In this work, we address these gaps by systematically investigating covariance descriptors for medical image classification. 
Our contributions are threefold:
\begin{itemize}
    \item To the best of our knowledge, we are the first to propose GVE-based covariance descriptors for image classification;
    \item We apply, for the first time, deep learning-based covariance descriptor classification with SPDNet \cite{huang2017riemannian} to medical images;
    \item We provide a comprehensive evaluation across conventional and learning-based classifiers, feature types, and diverse binary and multi-label classification tasks from the MedMNSIT benchmark \cite{yang2023medmnist}.
\end{itemize}
Our code is publicly available at \href{https://github.com/compai-lab/2026-isbi-mayr}{https://github.com/compai-lab/2026-isbi-mayr}.

\section{Methods}
\label{sec:methods}
The proposed method comprises three stages (see Fig. \ref{fig:overview}): (i) Feature extraction from medical images, (ii) computation of covariance descriptors, and (iii) classification of the descriptors using methods adapted to the SPD manifold. Each stage is described in detail below.

\subsection{Extraction of Handcrafted and General Vision Encoder Features}

\begin{figure}[t]
    \centering
    \includegraphics[width=\linewidth]{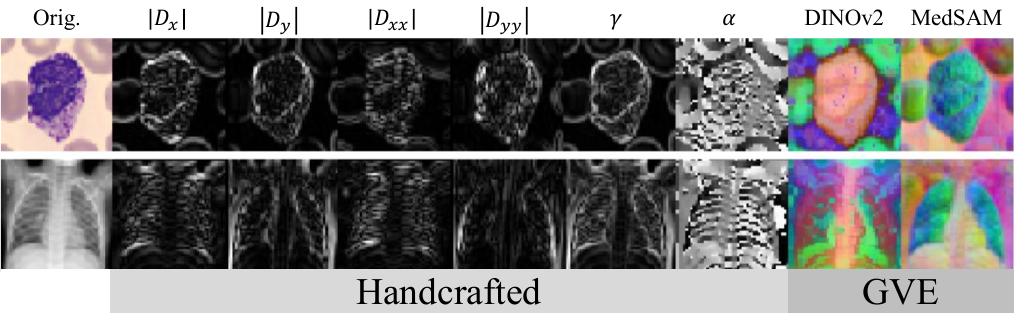}
    \caption{Handcrafted (HC) features capture local edge and gradient properties while features extracted from pretrained general vision encoders (GVEs) capture high-level semantics. Here, the first three principal components are shown for the GVE features.}
    \label{fig:method.conventional_vs_gve}
\end{figure}

\noindent\textit{Handcrafted image (HC) features}: HC features capture local image properties such as gradients and edges between neighboring pixels.

Following prior work \cite{avaznia2017breast}, we use 
\begin{equation}
    \mathbb{F}_{\mathrm{HC}} = (x, y, |D_x|, |D_y|, |D_{xx}|, |D_{yy}|, \lambda, \alpha)^T\in\mathbb{R}^{8},
\end{equation}
where $x, y\in [0, 1]$ denote normalized pixel coordinates, $D_x, D_y$ and $D_{xx}, D_{yy}$ are first- and second-order Sobel-filter based spatial derivatives in horizontal and vertical direction, $\gamma=\|(D_x, D_y)^T\|_2$ is the gradient magnitude, and $\alpha=\arctan(D_y/D_x)$ the gradient orientation.\\

\noindent\textit{General vision encoder (GVE) features}: 
In contrast, GVE features capture high-level semantics, long-range dependencies, and contextual relationships within the image.
We consider two pre-trained models: (i) DINOv2 \cite{oquab2023dinov2}, trained on large-scale natural images for general-purpose representation learning, and (ii) MedSAM \cite{MedSAM}, a domain-specific adaptation of the Segment Anything Model (SAM) \cite{kirillov2023segment} that is fine-tuned for medical image segmentation. 
We use the large variant of the DINOv2 model from which we extract 1024-dimensional features $\mathbb{F}_{\mathrm{DINOv2}}$. From the MedSAM model, the extracted features $\mathbb{F}_{\mathrm{MedSAM}}$ are 256-dimensional.
No additional fine-tuning is performed, and input images scaled to size 518x518 (DINOv2) and 1024x1024 (MedSAM) are processed into feature maps of size 37x37 and 64x64, respectively.
Fig. \ref{fig:method.conventional_vs_gve} illustrates HC and GVE features for two example images of the MedMNIST benchmark \cite{yang2023medmnist}.

\subsection{Covariance Descriptor Computation}
Covariance descriptors capture second-order statistics of feature vectors across all pixels.
Given the feature image $\mathbf{F}$ of size $C\times H\times W$
of an image, the covariance descriptor is the sample covariance matrix:
\begin{equation}
    \mathbf{C}=\frac{1}{WH - 1}\sum_{p\in [H]\times [W]} (\mathbf{F}_p - \bm{\mu})(\mathbf{F}_p-\bm{\mu})^T
\end{equation}
where $\bm{\mu}$ is the mean of the features in $\mathbf{F}$.

By construction, $\mathbf{C}$ is a symmetric positive definite (SPD) matrix.
Importantly, the set of SPD matrices forms a Riemannian manifold, a mathematical structure that describes the geometry of non-Euclidean spaces.
Since the SPD manifold does not follow Euclidean geometry, standard Euclidean metrics cannot accurately capture distances between SPD matrices. Instead, geometry-aware distance measures should be used. 
In this work, we employ the log-Euclidean metric (LEM) \cite{Arsigny2005}. For two SPD matrices $\mathbf{A},\mathbf{B}$ it is defined as: 
\begin{equation}
\label{eq:lem}
    d_{\mathrm{LEM}}(\mathbf{A}, \mathbf{B}) = \left\| \log(\mathbf{A}) - \log(\mathbf{B}) \right\|_F,
\end{equation} where $\log(\cdot)$ is the matrix logarithm and $||\cdot ||_F$ the Frobenius norm.
To compute the mean of a set of SPD matrices $\{\mathbf{A}_i\}$, we use the Karcher (Fr\'echet) mean with the LEM:
\begin{equation}
    \bar{\mathbf{A}} = \argmin_{\mathbf{A}} \sum_{i=1}^N d_{\mathrm{LEM}}^2(\mathbf{A}, \mathbf{A}_i).
\end{equation}
Locally, the SPD manifold can be approximated by a Euclidean tangent space, where conventional Euclidean methods apply. However, note that it cannot retain the geometric properties of the manifold. Mapping between the manifold and the local tangent space is performed with the Exponential and Logarithmic maps.
For more details on Riemannian manifolds, see \cite{Lee2012_Smo}.

\subsection{Classification methods for SPD matrices}
We evaluate two traditional and one learning-based classification method, 
each adapted to the Riemannian geometry of the SPD matrices:

\begin{itemize}
    \item \noindent\textit{Minimum Distance to Riemannian Mean (MDRM):} First, the covariance descriptors are obtained for each training image. Then, for each class, the Karcher mean of the corresponding descriptors is computed and used as a class descriptor. The test images are assigned to the class having the nearest class descriptor.
    \item \noindent\textit{Tangent Space Linear Discriminant Analysis (TSLDA):} Standard linear discriminant analysis is applied in the Euclidean tangent space at the Karcher mean of the training set, similar to
\cite{avaznia2017breast, talebi2019bloodcells}.
To avoid numerical issues, tangent vectors containing values above $10^{11}$ were discarded.
\item \noindent\textit{SPDNet:}
SPDNet \cite{huang2017riemannian} is a supervised deep network designed specifically for SPD matrices, originally proposed for action and face recognition. Its manifold-aware layers reduce dimensionality via learnable rotation and projection while preserving Riemannian structure. To the best of our knowledge, this approach has not been previously applied to medical imaging. 
\end{itemize}

\section{Experiments and Results}
\label{sec:results}
\subsection{Datasets and Implementation Details}

We evaluate our method on 11 single-target binary and multi-class datasets of the publicly available MedMNIST collection \cite{yang2023medmnist}. We use the datasets with images of size 64$\times$64.
The datasets cover a broad range of anatomies (e.g., colon, chest, breast, abdomen) and imaging modalities (e.g., x-ray, pathology, ultrasound, CT).
Note that most of these datasets are highly unbalanced. 
For the conventional methods, the GVE features are reduced to 16 dimensions using principal component analysis (PCA). For SPDNet, no such preprocessing was performed since dimensionality reduction is an intrinsic capability of SPDNet. To ensure a sufficient number of features for the handcrafted feature set, we extract 16 overlapping windows per image with a window size of 21 and a stride of 12. The windows are concatenated along the feature axis.
All Riemannian operations are implemented with the GeomStats framework \cite{miolane2020geomstats} and we build upon the PyTorch SPDNet implementation of \cite{Schwander_2020}. 
For training, we employ the manifold-adapted Adam optimizer \cite{kingma2014adam} of \cite{brooks2019second} and use a cross-entropy loss compensating the class imbalance via weighting. 
Models are trained for 250 epochs with early stopping, except for the large-scale datasets (\textit{Path}, \textit{OCT}, \textit{Tissue}), training is limited to 20 epochs due to computational constraints.
For the small datasets (\textit{Breast}, \textit{Retina}, \textit{Pneumonia}, \textit{Derma}, \textit{Blood}, \textit{OrganC}, and \textit{OrganS}), results are averaged across five random model initializations, and for the large datasets above, a single run is reported.

\begin{table*}[h]
	\centering
    \caption{Performance comparison of handcrafted (HC) features vs. learned GVE features (MedSAM and DINOv2) for traditional and learning based covariance descriptor classification. Balanced accuracy is reported. The \colorbox{gray!25}{best feature type} per classification method and the \textbf{best configuration} per dataset are highlighted.}
	\input{tab_ablation}
	\label{tab:ablation}
\end{table*}

\begin{table*}[h]
	\centering
    \caption{Performance comparison of our method to state-of-the-art results reported in the literature in terms of AUC. The \textbf{best} and \underline{second best} results per dataset are highlighted. (Pneu., A, C, and S stands for Pneumonia, Axial, Coronal, and Sagittal.)}
	\input{tab_main}

	\label{tab:main}
\end{table*}

\subsection{Experimental Evaluation}
Our evaluation is structured in two stages. 
In the first stage, we analyze the effect of different feature representations and classification methods on covariance descriptor performance. 
In the second stage, we compare our best-performing configuration to established baselines on MedMNIST.

\subsubsection{Comparison across feature types and classifiers}
\label{sec:findings1}
We first compare covariance descriptors based on handcrafted features (HC) with such based on learned features from GVEs across the three classifiers: MDRM, TSLDA, and SPDNet.
The balanced accuracy results reported in Tab.~\ref{tab:ablation} reveal several findings:

\textit{1) Learned GVE features outperform HC features:} 
The results show that across all classifiers, covariance descriptors constructed from learned features consistently outperform those from HC features, often by a substantial margin. 
This indicates that medical image classification benefits from richer feature representations. 
HC features capture only local intensity gradients and positional information, which limits their ability to represent complex anatomical structures. In contrast, GVE features encode higher-level spatial dependencies and semantic context that captures more discriminative relational information.
The difference in feature expressiveness is also reflected in Fig. 2.
The performance improvement is particularly evident in datasets with texture or shape patterns, such as \textit{Derma}, \textit{OrganA}, \textit{OrganC}, \textit{OrganS}, where handcrafted features fail to account for global structural cues.

\textit{2) DINOv2-derived features outperform MedSAM features:}
Among the two GVEs evaluated, DINOv2 features lead to higher balanced accuracy than MedSAM features across nearly all datasets and classifiers. 
This is surprising, given that MedSAM is fine-tuned on medical images, while DINOv2 is trained on natural images only. A reason for this could be that MedSAM is optimized for segmentation rather than classification and emphasizes boundary localization, which may limit the discriminative power of its features.

\textit{3) SPDNet combined with DINOv2 features achieves the best overall performance:} 
SPDNet consistently outperforms both MDRM and TSLDA across all datasets.
It learns to project covariance descriptors into lower-dimensional tangent spaces while preserving the geometric structure of the descriptors.
The advantage of SPDNet is most prominent when applied to covariance descriptors derived from DINOv2 features. This configuration outperforms all others in every single case, and for \textit{OCT}, \textit{Retina}, \textit{Derma}, and \textit{Tissue}, the gap between MedSAM and DINOv2 features is particularly large. 

These results demonstrate the benefit of leveraging pre-trained general-purpose feature encoders for manifold-aware covariance descriptor learning.

\subsubsection{Comparison with Baseline Methods} 
In the second stage of evaluation, we compare our best-performing configuration (SPDNet with DINOv2 features, see Tab. \ref{tab:ablation}) against state-of-the-art baseline architectures that have been applied to MedMNIST in the literature, including ResNet and ViT-based methods. Since these baselines report results in terms of AUC, we follow this standard for comparison. The results are summarized in Tab.~\ref{tab:main}.

It can be observed that our method achieves comparable or superior AUC performance in 7 out of the 11 datasets. 
This highlights that covariance descriptors based on pretrained GVE features are competitive and, in many cases, surpass standard CNN and transformer-based classification models trained end-to-end on MedMNIST.
Consistent with the previous findings of Sec. \ref{sec:findings1}, DINOv2-based descriptors consistently outperform MedSAM-based descriptors across all datasets, except for the \textit{OrganA} dataset, for which both perform on par. 
These results confirm that covariance descriptors are a powerful representation across diverse anatomies and modalities, where SPDNet forms a promising alternative to state-of-the-art deep learning-based classification approaches.

\section{Conclusion}
In this work, we investigated covariance descriptors for medical image classification.
We evaluated handcrafted and learned features from two general vision encoders (GVEs) models across 11 diverse datasets from the MedMNIST benchmark, as well as three classification methods designed for SPD matrices, including the learning-based SPDNet. 
Our results show that GVE feature-based covariance descriptors outperform handcrafted ones.
Among the GVE models, DINOv2 consistently yielded the highest performance, highlighting its strong transferability to the medical domain.
Moreover, SPDNet outperformed traditional Riemannian classifiers across all feature types by learning manifold-aware dimensionality reductions while retaining meaningful feature correlations.
Overall, this work demonstrates the power of GVE-based covariance descriptor representations for medical image classification, particularly when used within manifold-aware deep learning. 
Future directions include extending this approach to 3D data and investigating multimodal covariance descriptors, e.g., for fusing imaging and clinical metadata.

\section{Compliance with ethical standards}
\label{sec:ethics}
This research study was conducted retrospectively using human subject data made available in open access by \cite{yang2023medmnist}. Ethical approval was not required, as confirmed by the license attached to the open-access data.

\section{Conflict of Interest}
The authors have no relevant financial or non-financial interests to disclose.

\section{Acknowledgements}
AR received funding from the Munich Center for Machine Learning.

\bibliographystyle{IEEEtran}
\bibliography{bib}

\end{document}

%% file: tab_ablation.tex
\begin{scriptsize}
\begin{tabularx}{0.8\linewidth}{lccc|ccc|ccc}
    \toprule
    & \multicolumn{6}{c|}{\textit{Traditional methods}} & \multicolumn{3}{c}{\textit{DL-based methods}} \\
    & \multicolumn{3}{c}{\textbf{MDRM}} & \multicolumn{3}{c|}{\textbf{TSLDA}} & \multicolumn{3}{c}{\textbf{SPDNet}} \\
  
    \midrule
    & $\mathbb{F}_{\mathrm{HC}}$ & $\mathbb{F}_{\mathrm{MedSAM}}$ & $\mathbb{F}_{\mathrm{DINOv2}}$ & $\mathbb{F}_{\mathrm{HC}}$ & $\mathbb{F}_{\mathrm{MedSAM}}$ & $\mathbb{F}_{\mathrm{DINOv2}}$ & $\mathbb{F}_{\mathrm{HC}}$ & $\mathbb{F}_{\mathrm{MedSAM}}$ & $\mathbb{F}_{\mathrm{DINOv2}}$ \\
    &  &  &  & & & & & \textit{(ours)} & \textit{(ours)} \\
     
    \cline{2-10}
    Path      & 45.6                     & 69.0                         & \cellcolor{gray!25}79.9      & 18.2                     & 11.5                         & \cellcolor{gray!25}21.9      & 64.9 & 88.5 & \cellcolor{gray!25}\textbf{90.2}\\
    Derma     & 26.2                     & \cellcolor{gray!25}38.6      & 34.1                         & 14.7                     & \cellcolor{gray!25}28.1      & 14.4                         & 41.6 & 56.3 & \cellcolor{gray!25}\textbf{72.7}\\
    OCT       & 48.5                     & 42.1                         & \cellcolor{gray!25}55.7      & 25.0                     & \cellcolor{gray!25}46.1      & 25.0                         & 67.5 & 75.3 & \cellcolor{gray!25}\textbf{87.3}\\
    Pneumonia & \cellcolor{gray!25}83.1  & 79.4                         & 73.8                         & 63.5                     & \cellcolor{gray!25}83.9      & 83.5                         & 81.3 & 87.8 & \cellcolor{gray!25}\textbf{91.9}\\
    Retina    & 26.3                     & 34.2                         & \cellcolor{gray!25}39.3      & 27.0                     & 35.4                         & \cellcolor{gray!25}41.9      & 36.9 & 41.6 & \cellcolor{gray!25}\textbf{51.6}\\
    Breast    & 72.5                     & \cellcolor{gray!25}73.6      & 71.2                         & \cellcolor{gray!25}71.6  & 68.8                         & 55.7                         & 71.1 & 80.5 & \cellcolor{gray!25}\textbf{85.3}\\
    Blood     & 49.9                     & 76.9                         & \cellcolor{gray!25}78.3      & 53.7                     & 81.5                         & \cellcolor{gray!25}85.4      & 83.2 & 96.1 & \cellcolor{gray!25}\textbf{98.7}\\
    Tissue    & 21.5                     & 31.1                         & \cellcolor{gray!25}33.6      & 12.5                     & \cellcolor{gray!25}35.5      & 17.3                         & 33.8 & 50.9 & \cellcolor{gray!25}\textbf{56.3}\\
    OrganA    & 42.7                     & 72.0                         & \cellcolor{gray!25}75.8      & 62.1                     & \cellcolor{gray!25}75.4      & 10.7                         & 58.9 & 90.7 & \cellcolor{gray!25}\textbf{93.8}\\
    OrganC    & 39.1                     & \cellcolor{gray!25}72.1      & 70.4                         & 64.4                     & \cellcolor{gray!25}74.2      & 9.4                          & 64.0 & 89.6 & \cellcolor{gray!25}\textbf{91.5}\\
    OrganS    & 34.9                     & 54.1                         & \cellcolor{gray!25}58.1      & 49.5                     & \cellcolor{gray!25}58.5      & 10.7                         & 46.4 & 73.2 & \cellcolor{gray!25}\textbf{76.3}\\
      \bottomrule  
\end{tabularx}
\end{scriptsize}

%% file: tab_main.tex
 \begin{scriptsize}
 \begin{tabularx}{0.8\linewidth}{lllllllllllll}
     \toprule
                                                          &  Path            & Derma            & OCT              & Pneu.            & Retina           & Breast           & Blood            & Tissue           & \multicolumn{3}{c}{Organ} \\
                                                          &                  &                  &                  &                  &                  &                  &                  &                  & A                & C                & S                 \\
     \midrule
     ResNet-18 \cite{yang2023medmnist}                    & 98.9             & 92.0             & 95.8             & 95.6             & 71.0             & 89.1             & \underline{99.8} & 93.3             & \textbf{99.8}    & \underline{99.4} & 97.4              \\
     ResNet-50 \cite{yang2023medmnist}                    & 98.9             & 91.2             & 95.8             & 96.2             & 71.6             & 86.6             & 99.7             & 93.2             & \textbf{99.8}    & 99.3             & 97.5              \\
     AutoML \cite{yang2023medmnist}                       & 94.4             & 91.4             & \underline{96.3} & 99.1             & 75.0             & 91.9             & \underline{99.8} & 92.4             & 99.0             & 98.8             & 96.4              \\
     MedViT-T \cite{manzari2023medvit}                    & \underline{99.4} & 91.4             & 96.1             & \underline{99.3} & 75.2             & \underline{93.4} & 99.6             & \underline{94.3} & 99.5             & 99.1             & 97.2              \\
     MedViT-S \cite{manzari2023medvit}                    & 99.3             & \underline{93.7} & 96.0             & \textbf{99.5}    & \underline{77.3} & \textbf{93.8}    & 99.7             & \textbf{95.2}    & 99.6             & 99.3             & \underline{98.7}  \\
     MedViT-L \cite{manzari2023medvit}                    & 98.4             & 92.0             & 94.5             & 99.1             & 75.4             & 92.9             & 99.6             & 93.5             & \underline{99.7} & \underline{99.4} & 97.3              \\
     \midrule                                                                
     \multicolumn{1}{l}{SPDNet + MedSAM \textit{(ours)}}  & \underline{99.4} & 88.5             & 95.4             & 98.0             & 75.7             & 89.9             & \underline{99.8} & 88.1             & 99.5             & 99.3             & 97.3              \\
     \multicolumn{1}{l}{SPDNet + DINOv2 \textit{(ours)}}  & \textbf{99.5}    & \textbf{95.5}    & \textbf{98.4}    & 99.2             & \textbf{83.7}    & 91.4             & \textbf{99.9}    & 90.7             & 99.5             & \textbf{99.8}    & \textbf{99.6}     \\
   
     \bottomrule
 \end{tabularx}
 \end{scriptsize}

%% file: bib.bib
@inproceedings{huang2017riemannian,
   title={A {Riemannian} Network for {SPD} Matrix Learning},
   volume={31},
   DOI={10.1609/aaai.v31i1.10866},
   number={1},
   journal={Proceedings of the AAAI Conference on Artificial Intelligence},
   booktitle={2017: The Thirty-First AAAI Conference on Artificial Intelligence},
   author={Huang,
   Zhiwu and Van Gool,
   Luc},
   year={2017},
   month={Feb.}
}

@article{oquab2023dinov2,
  title={{DINOv2}: Learning Robust Visual Features without Supervision},
  author={Oquab, Maxime and Darcet, Timoth{\'e}e and Moutakanni, Th{\'e}o and Vo, Huy and Szafraniec, Marc and Khalidov, Vasil and Fernandez, Pierre and Haziza, Daniel and Massa, Francisco and El-Nouby, Alaaeldin and others},
  journal={arXiv preprint arXiv:2304.07193},
  year={2023}
}

@article{MedSAM,
  title={Segment anything in medical images},
  author={Ma, Jun and He, Yuting and Li, Feifei and Han, Lin and You, Chenyu and Wang, Bo},
  journal={Nature Communications},
  volume={15},
  pages={654},
  year={2024}
}

@article{yang2023medmnist,
  title={{MedMNIST} v2 -- A large-scale lightweight benchmark for 2{D} and 3{D} biomedical image classification},
  author={Yang, Jiancheng and Shi, Rui and Wei, Donglai and Liu, Zequan and Zhao, Lin and Ke, Bilian and Pfister, Hanspeter and Ni, Bingbing},
  journal={Scientific Data},
  volume={10},
  number={1},
  pages={41},
  year={2023},
  publisher={Nature Publishing Group UK London}
}

@inproceedings{tuzel2006region,
  title={Region Covariance: A Fast Descriptor for Detection and Classification},
  author={Tuzel, Oncel and Porikli, Fatih and Meer, Peter},
  editor={Leonardis, Ale{\v{s}} and Bischof, Horst and Pinz, Axel},
  booktitle={Computer Vision -- ECCV 2006},
  pages={589--600},
  year={2006},
  organization={Springer},
  publisher={Springer},
  isbn={978-3-540-33835-2}
}

@InProceedings{Tabia_2014_CVPR,
  author = {Tabia, Hedi and Laga, Hamid and Picard, David and Gosselin, Philippe-Henri},
  title = {Covariance Descriptors for 3{D} Shape Matching and Retrieval},
  booktitle = {2014 IEEE Conference on Computer Vision and Pattern Recognition},
  month = {June},
  year = {2014}
}

@inproceedings{sanin2013spatio,
  title={Spatio-temporal covariance descriptors for action and gesture recognition},
  author={Sanin, Andres and Sanderson, Conrad and Harandi, Mehrtash T and Lovell, Brian C},
  booktitle={2013 IEEE Workshop on applications of Computer Vision (WACV)},
  pages={103--110},
  year={2013},
  organization={IEEE}
}

@inproceedings{kirillov2023segment,
  title={Segment Anything},
  author={Kirillov, Alexander and Mintun, Eric and Ravi, Nikhila and Mao, Hanzi and Rolland, Chloe and Gustafson, Laura and Xiao, Tete and Whitehead, Spencer and Berg, Alexander C. and Lo, Wan-Yen and Dollar, Piotr and Girshick, Ross},
  booktitle={Proceedings of the IEEE/CVF International Conference on Computer Vision (ICCV)},
  pages={4015--4026},
  year={2023}
}

@article{miolane2020geomstats,
  title={Geomstats: A {Python} package for {Riemannian} geometry in machine learning},
  author={Miolane, Nina and Guigui, Nicolas and Le Brigant, Alice and Mathe, Johan and Hou, Benjamin and Thanwerdas, Yann and Heyder, Stefan and Peltre, Olivier and Koep, Niklas and Zaatiti {et al.}, Hadi},
  journal={Journal of Machine Learning Research},
  volume={21},
  number={223},
  pages={1--9},
  year={2020}
}

@inproceedings{brooks2019second,
  title={Second-order networks in PyTorch},
  author={Brooks, Daniel and Schwander, Olivier and Barbaresco, Fr{\'e}d{\'e}ric and Schneider, Jean-Yves and Cord, Matthieu},
  booktitle={Geometric Science of Information},
  pages={751--758},
  year={2019},
  
  publisher={Springer},
  isbn={978-3-030-26980-7}
}

@article{kingma2014adam,
  title={Adam: A Method for Stochastic Optimization},
  author={Kingma, Diederik P and Ba, Jimmy},
  journal={arXiv preprint arXiv:1412.6980},
  year={2014}
}

@article{manzari2023medvit,
  title={{MedViT}: A robust vision transformer for generalized medical image classification},
  author={Manzari, Omid Nejati and Ahmadabadi, Hamid and Kashiani, Hossein and Shokouhi, Shahriar B and Ayatollahi, Ahmad},
  journal={Computers in Biology and Medicine},
  volume={157},
  pages={106791},
  year={2023},
  publisher={Elsevier}
}

@inproceedings{avaznia2017breast,
  title={Breast cancer classification using covariance description in {Riemannian} geometry},
  author={Avaznia, Cyrus and Naghavi, Seyyed Hamed and Menhaj, Mohammad Bagher and Talebi, Hamed},
  booktitle={2017 10th Iranian Conference on Machine Vision and Image Processing (MVIP)},
  pages={110--113},
  year={2017},
  organization={IEEE}
}

@Article{talebi2019bloodcells,
  author        = {Talebi, Hamed and Ranjbar, Amin and Davoudi, Alireza and Gholami, Hamed and Menhaj, Mohammad Bagher},
  title         = {High Accuracy Classification of White Blood Cells using {TSLDA} Classifier and Covariance Features},
  year          = {2019},
  month         = jun,
  archiveprefix = {arXiv},
  copyright     = {arXiv.org perpetual, non-exclusive license},
  doi           = {10.48550/ARXIV.1906.05131},
  eprint        = {1906.05131},
  primaryclass  = {cs.CV},
  publisher     = {arXiv},
  journal={arXiv preprint arXiv:1906.05131},
}

@inproceedings{dosovitskiy2020vit,title	= {An Image is Worth 16x16 Words: Transformers for Image Recognition at Scale},author	= {Alexander Kolesnikov and Alexey Dosovitskiy and Dirk Weissenborn and Georg Heigold and Jakob Uszkoreit and Lucas Beyer and Matthias Minderer and Mostafa Dehghani and Neil Houlsby and Sylvain Gelly and Thomas Unterthiner and Xiaohua Zhai},year	= {2021}, booktitle = {International Conference on Learning Representations (ICLR)}}

@InProceedings{Cirujeda2015_3D,
  title = {3{D} {Riesz}-wavelet based Covariance descriptors for texture classification of lung nodule tissue in {CT}},
 author={Cirujeda, Pol and Müller, Henning and Rubin, Daniel and Aguilera, Todd A. and Loo, Billy W. and Diehn, Maximilian and Binefa, Xavier and Depeursinge, Adrien},

  booktitle = {2015 37th Annual International Conference of the IEEE Engineering in Medicine and Biology Society (EMBC)},
  year = {2015},
  pages = {7909-7912},
  doi = {10.1109/EMBC.2015.7320226}
}

@Article{Khan2015,
  title = {A Global Covariance Descriptor for Nuclear Atypia Scoring in Breast Histopathology Images}, 
  author = {Khan, Adnan Mujahid and Sirinukunwattana, Korsuk and Rajpoot, Nasir},
  journal = {IEEE Journal of Biomedical and Health Informatics}, 
  year = {2015},
  volume = {19},
  number = {5},
  pages = {1637-1647},
  doi = {10.1109/JBHI.2015.2447008}
}

@InProceedings{Stanitsas2016,
  title = {Evaluation of feature descriptors for cancerous tissue recognition}, 
  author = {Stanitsas, Panagiotis and Cherian, Anoop and Xinyan Li and Truskinovsky, Alexander and Morellas, Vassilios and Papanikolopoulos, Nikolaos},
  booktitle = {2016 23rd International Conference on Pattern Recognition (ICPR)}, 
  year = {2016},
  volume = {},
  number = {},
  pages = {1490-1495},
  doi = {10.1109/ICPR.2016.7899848}
}

@Article{Ahmadi2024,
  title = {Ricci flow-based brain surface covariance descriptors for diagnosing Alzheimer’s disease},
  journal = {Biomedical Signal Processing and Control},
  volume = {93},
  pages = {106212},
  year = {2024},
  issn = {1746-8094},
  doi = {https://doi.org/10.1016/j.bspc.2024.106212},
  author = {Fatemeh Ahmadi and Mohamad-Ebrahim Shiri and Behroz Bidabad and Maral Sedaghat and Pooran Memari},
}

@incollection{Arsigny2005,
  author    = {Arsigny, Vincent and Fillard, Pierre and Pennec, Xavier and Ayache, Nicholas},
  pages     = {115--122},
  publisher = {Springer},
  title     = {Fast and Simple Calculus on Tensors in the Log-{Euclidean} Framework},
  year      = {2005},
  isbn      = {9783540320944},
  booktitle = {Medical Image Computing and Computer-Assisted Intervention – MICCAI 2005},
  doi       = {10.1007/11566465_15},
  issn      = {1611-3349},
}

@Book{Lee2012_Smo,
  author    = {Lee, John M.},
  publisher = {Springer},
  title     = {Introduction to Smooth Manifolds},
  year      = {2012},
  isbn      = {9781441999825},
  doi       = {10.1007/978-1-4419-9982-5},
  issn      = {0072-5285},
  journal   = {Graduate Texts in Mathematics},
}

@misc{Schwander_2020, 
title={torchspdnet}, 
url={https://gitlab.lip6.fr/schwander/torchspdnet}, 
journal={GitLab}, 
author={Schwander, Olivier}, 
year={2020}, 
month={Jun},
}
